\documentclass[runningheads,a4paper]{llncs}

\usepackage{graphicx}
\usepackage{amsmath}
\usepackage{amssymb}

\usepackage{array}
\usepackage{multirow}
\usepackage{tikz}
\usetikzlibrary{positioning,arrows}

\usepackage{cite}

\usepackage[capitalise,nameinlink]{cleveref}
\crefname{section}{Sect.}{Sect.}
\Crefname{section}{Section}{Sections}

\begin{document}

\title{Max-Cosine Matching Based Neural Models for Recognizing Textual Entailment}
\titlerunning{Max-Cosine Matching Based Neural Models for RTE}

\author{Zhipeng Xie \and Junfeng Hu}
\institute{Shanghai Key Laboratory of Data Science\\
School of Computer Science, Fudan University, China\\
\email{\{xiezp, 15210240075\}@fudan.edu.cn}
}
			
\maketitle

\begin{abstract}
Recognizing textual entailment is a fundamental task in a variety of text mining or natural language processing applications. This paper proposes a simple neural model for RTE problem. It first matches each word in the hypothesis with its most-similar word in the premise, producing an augmented representation of the hypothesis conditioned on the premise as a sequence of word pairs. The LSTM model is then used to model this augmented sequence, and the final output from the LSTM is fed into a softmax layer to make the prediction. Besides the base model, in order to enhance its performance, we also proposed three techniques: the integration of multiple word-embedding library, bi-way integration, and ensemble based on model averaging. Experimental results on the SNLI dataset have shown that the three techniques are effective in boosting the predicative accuracy and that our method outperforms several state-of-the-state ones.
\end{abstract}


\section{Introduction}\label{sec:intro}

In natural language text, there are always different ways to express the same meaning. This surface-level variability of semantic expressions is fundamental in tasks related to natural language processing and text mining. Textual entailment recognition (or RTE in short) is a specific semantic inference approach to model surface-level variability. As formulated by Dagan and Glickman \cite{rte}, the task of Recognizing Textual Entailment is to decide whether the meaning of a text fragment $Y$ (called the Hypothesis) can be inferred (is inferred) from another text fragment $X$ (called the Premise). Giampiccolo et al. \cite{rte-extended} extended the task to include the additional requirement that systems identify when the Hypothesis contradicts the Premise. The semantic inference needs are pervasive in a variety of NLP or text mining applications \cite{rte-rational}, inclusive of but not limited to, question-answering \cite{entailment-qa}, text summarization \cite{entailment-ts}, and information extraction. Given a pair of premise $X$ and hypothesis $Y$, the relation between them may be: Entailment ($Y$ can be inferred from $X$), Contradiction ($Y$ is inferred to contradict $X$), or Neutral ($X$ and $Y$ are unrelated to each other). \Cref{tab:nli-example} presents a simple example to illustrate these three relations.

\begin{table}[t]
\caption{An illustrative RTE example}
\label{tab:nli-example}
\begin{center}
\begin{tabular}{||>{\centering\arraybackslash}m{3.5cm} |>{\centering\arraybackslash}m{4.8cm} |>{\centering\arraybackslash}m{2cm} || }
\hline
	\bf Premise        & \bf Hypothesis    &  \bf Relation \\ 
\hline\hline
  & Giving money to a poor man has good consequences.  & Entailment \\
If you help the needy, God will reward you.	                        & Giving money to a poor man has no consequences.           & Contradiction  \\
	                        & Giving money to a poor man will make you a better person. & Neutral \\ 
\hline
\end{tabular}
\end{center}
\end{table}

A lot of research work has been devoted to the RTE problem in the last decade. The mainstream methods for recognizing textual entailment can be roughly divided into two categories:
\begin{itemize}
\item The first category attempts to provide a sequence of transformations allowing to derive the hypothesis $Y$ from the premise $X$, by applying one transformation rule at each step. 
\item The second category simply thinks of RTE problem as a classification problem, where features (manually defined or automatically constructed) are extracted from the premise-hypothesis pairs. 
\end{itemize}

In transformation-based RTE methods (also called rule-based methods), the underlying idea is to make use of inference rules (or entailment rules) for making transformation. However, the lack of such knowledge has been a major obstacle to improving the performance on RTE problem. The acquisition of entailment rules can be done either by learning algorithms which extract entailment rules from large text corpora, or by methods which extract rules from manually constructed knowledge resources.

Some research works have focused on extraction of entailment rules from manually constructed knowledge resources. WordNet \cite{wordnet} is the most prominent resource to extract entailment rules from. The synonymy and hypernymy relations (called substitutable relations) can be exploited to do direct substitution. To make use of the other non-substitutable relations (such as entailment and cause relations), Szpektor and Dagan \cite{rte-wordnet-augment} populated these non-substitutable relations with argument mapping which are extracted various resource, and thus extended WordNet's inferential relations at the syntactic representation level. FrameNet \cite{framenet} is another manually constructed lexical knowledge base for entailment rule extraction. Aharon et al. \cite{rte-framenet} detected the entailment relations implied in FrameNet, and utilized FrameNet's annotated sentences and relations between frames to extract both the entailment relations and their argument mappings.

Although the entailment rules extracted from manually constructed knowledge resources have achieved sufficiently accuracy, their coverage is usually severely limited. A lot of research work has been devoted to learning entailment rules from a given text corpus. The DIRT algorithm proposed by Lin and Pantel \cite{dirt} was based on the so-called Extended Distributional Hypothesis which states that phrases occurring in similar contexts are similar. An inference rule extracted by DIRT algorithm is actually a paraphrase rule, which is a pair of language patterns that can replace each other in a sentence. In DIRT, the language patterns are chains in dependency trees, with placeholders for nouns at the end of this chain. Different from the Extended Distributional Hypothesis adopted by DIRT, Glickman and Dagan \cite{instance-paraphrase} proposed an instance-based approach, which uses linguistic filters to identify paraphrase instances that describe the same fact and then rank the candidate paraphrases based on a probabilistically motivated paraphrase likelihood measure. Sekine \cite{link-paraphrase} extracted the phrase between two named entities as candidate linear pattern, then identified a keyword in each phrase and joined phrases with the same keyword into sets, and finally linked sets which involve the same pairs of individual named entities. The sets or the links can be treated as paraphrases.

Besides paraphrase rules (which can be thought of as a specific case of entailment rules), a more general notion needed for RTE is that of entailment rules \cite{rte}. An entailment rule is a directional relation between two language patterns, where the meaning of one can be entailed or inferred from the meaning of the other, but not vice versa. Pekar \cite{verb-rte} proposed a three-step method: it first identifies pairs of discourse-related clauses, and then creates patterns by extracting pairs of verbs along with relevant information as to their syntactic behaviour, and finally scores each verb pair in terms of plausibility of entailment. Recently, Szpektor et al. \cite{unsupervised-rte} presented a fully unsupervised learning algorithm for Web-based extraction of entailment rules. The algorithm takes as its input a lexical-syntactic template and searches the Web for syntactic templates that participate in an entailment relation with the input template.

Recently, with the availability of large high-quality dataset, especially with the Stanford Natural Language Inference (SNLI) corpus published in 2015 by Bowman et al. \cite{snli-dataset}, there comes an upsurge of end-to-end neural models for RTE, where the fundamental problem is how to model a sentence pair $(X,Y)$. The first and simplest model, proposed by Bowman et al. \cite{snli-dataset}, encodes the premise and the hypothesis with two separate LSTMs, and then feeds the concatenation of their final outputs into a MLP for classification. This model does not take the interaction between the premise and the hypothesis into consideration. Several follow-ups have been proposed to solve this problem by modeling their interaction with a variety of attentive mechanisms \cite{attention-lstm}\cite{match-lstm}\cite{intra-attention}. These models treat sentences as word sequences, but some others adopt more principled choice to work on the tree-structured sentences, by explicitly model the compositionality and the recursive structure of natural language over trees. Such kind of work includes the Stack-augmented Parser-Interpreter Neural Network (SPINN) \cite{spinn} and Tree-based Convolutional Neural Network (TBCNN) \cite{tbcnn}. In the section \ref{sec:related}, we shall have a look at all these neural models in more detail.

In this paper, we propose a simple neural method, called MaxConsineLSTM,based on max-cosine matching for natural language inference. It first matches each word in the hypothesis (or the premise) with its most-similar word in the premise (or the hypothesis), and obtains a representation of hypothesis (or the premise conditioned on the premise (or the hypothesis). Then, LSTM is used to model the enhanced representations of hypothesis and premise into dense vectors. And finally, we concatenate the two dense vectors and feed it into a softmax layer to make the final decision about the relation between them. Experimental results have shown that our method achieves better or comparable performance when compared with state-of-the-art methods.

\section{Related Work}\label{sec:related}

In this section, we review some neural models that work for recognizing textual entailment.

The first and simplest neural model to RTE was proposed by Bowman et al. \cite{snli-dataset} in 2015, which uses separate LSTMs \cite{lstm} to encode the premise and the hypothesis as dense fixed-length vectors and then feeds their concatenation into a multi-layer perceptron (MLP) or other classifiers for classification. It learns the sentence representation of premise and hypothesis independently, and does not take their interaction into consideration.

This first neural model suffer from the fact that the hypothesis and the premise are modeled independently, and thus the information cannot flow between them. To solve this problem, a sequential LSTM model is proposed in \cite{attention-lstm}. An LSTM reads the premise, and a second LSTM with different parameters reads a delimiter and the hypothesis, but its memory state is initialized with the final cell state of the previous LSTM. In this way, information from the premise can flow to the encoding of hypothesis.

To further facilitate the information flow between the premise and the hypothesis, Rockt\"aschel et al. \cite{attention-lstm} applied a neural attention model which can achieve better performance. When the second LSTM processes the hypothesis one word at a time, the first LSTM's output vectors are attended over, generating attention weights over all output vectors of the premise for every word in the hypothesis. The final sentence-pair representation is obtained from the last attention-weighted representation of the premise and the last output vector of the hypothesis. It outperforms Bowman et al. (2015) in that it checks for entailment or contradiction of individual word- and phrase-pairs. 

Wang and Jiang \cite{match-lstm} used an LSTM to perform word-by-word matching of the hypothesis with the premise. It is expected that the matching results that are critical for the final prediction will be ``remembered'' by the LSTM while less important matching results will be ``forgotten''. 

The attentive mechanisms used in \cite{attention-lstm} and \cite{match-lstm} are both between the hypothesis and the premise. It is sometimes helpful to exploit the attentive mechanism within the hypothesis or the premise. the long short-term memory-network (LSTMN) proposed by Cheng et al. \cite{intra-attention} induces undirected relations among tokens as an intermediate step of learning representations, which can be thought of as an intra-attention mechanism. It has also been manifested that the intra-attention mechanism can lead to representations of higher quality.

The algorithms described above all deal with sentences as sequences of word vectors, and learn sequence-based recurrent neural networks to map them to sentence vectors. Another more principled choice is to learn the tree-structured recursive networks. Recursive neural networks explicitly model the compositionality and the recursive structure of natural language over tree. Bowman et al. \cite{spinn} introduced the Stack-augmented Parser-Interpreter Neural Network (or SPINN in short) to combine parsing and interpretation within a single tree-sequence hybrid model by integrating tree-structured sentence interpretation into the linear sequential structure of a shift-reduce parser. 

Mou et al. \cite{tbcnn} proposed a tree-based convolutional neural network (TBCNN) to capture sentence-level semantics. TBCNN is more robust than sequential convolution in terms of word order distortion introduced by determinators, modifiers, etc. In TBCNN, a pooling layer aggregates information along tree, serving as a way of semantic compositionality. Finally, two sentences' information is combined and fed into a softmax layer for output.

\section{Base Method}\label{sec:methods}

Recognizing textual entailment is concerned about the relation between two sequences - the premise $X$ and the hypothesis $Y$. The commonly used encoder-decoder architecture processes the second sequence conditioned on the first one.

\begin{figure}
\centering
\pgfmathsetseed{1}
\begin{tikzpicture}[rounded corners=2pt, scale=0.7]
    \foreach \t in {-5,...,0} {
    \pgfmathtruncatemacro{\addsix}{\t + 6};
    \node[fill=blue!60!white] (prem\addsix) at (\t, 0) {$\mathbf{x}_\addsix$};
   }
   \foreach \t in {1,...,4} {
    	\pgfmathtruncatemacro{\prem}{-random(0, 5)};
    	\pgfmathtruncatemacro{\addsix}{\prem + 6}; 
    	\node[rectangle, fill=yellow!80!black, text width=.4cm] (hypo\t) at (\t * 2, 0) {$\mathbf{y}_\t$};

	    \draw[densely dotted, <-,semithick, blue!70!black] (prem\addsix) edge[out=315, in=200] node[sloped] {\small{$\gamma$}} (hypo\t) ;
      
    \node[draw, outer sep=2, rectangle, fill=green!20] (step\t) [above=of hypo\t] {$\mathbf{c}_\t^h$};
    \node[outer sep=2, rectangle, fill=orange!90!black] (hid\t) [above=of step\t] {$\mathbf{h}_\t$};
    \path[->] (hypo\t) edge (step\t);
    \path[->] (step\t) edge (hid\t);
  }
        
  \node[draw, outer sep=2, rectangle, fill=red!50] (softmax) [above=of hid4] {Softmax Layer};
  \path[->] (hid4) edge (softmax);
  \foreach \t in {1,...,3} {
      \pgfmathtruncatemacro{\next}{\t + 1};
    \path[->] (step\t) edge node[above]{$\mathbf{h}_\t$} (step\next);
  }
\end{tikzpicture}

\caption{Architecture of MaxCosine-LSTM}
\label{fig:maxcosinelstm}
\end{figure}
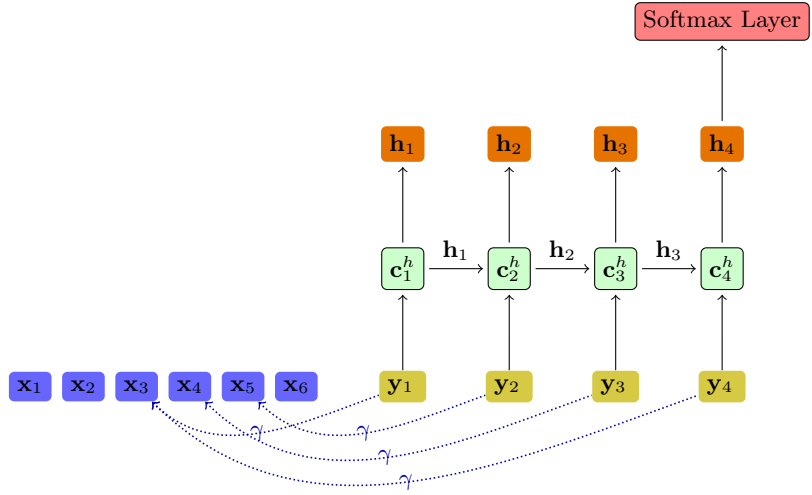

In this paper, we establish the connection of the hypothesis to the premise at the word level, where each word in the hypothesis is matched to and paired with its most-similar word in the premise. It leads to a simple base method called $MaxCosineLSTM$, consisting of three steps as illustrated in Figure \ref{fig:maxcosinelstm}:
\begin{itemize}
\item \textbf{Step 1: }(Word Matching) Each word $y_t$ in the hypothesis $Y$ is matched to its most-similar word (denoted as $\gamma(y_t)$ in $X$, where the similarity between two words is measured as the cosine similarity between their embeddings. Such a match strategy can be thought of as a conditional representation of the hypothesis on the given premise;
\item \textbf{Step 2: }(Sequence Modeling) For each time step $1\le t \le m$, the concatenation of the word embedding of $y_i$ and that of $\gamma(y_i)$ is fed into an LSTM layer, yield a vector representation of the hypothesis conditional on the premise;
\item \textbf{Step 3: }(Decision Making) The final output of the LSTM layer is fed into a softmax layer to get the final decision about the relation between $X$ and $Y$: Entailment, Contradiction, or Neutral.
\end{itemize}

\subsection{Word Matching with Cosine Similarity}

To judge whether a hypothesis $Y$ can be inferred from a given premise $X$, it is of importance to get to know whether each word in $Y$ expresses a similar meaning as one word in the premise $X$. Distributional Hypothesis proposed by Harris \cite{distributional-hypothesis} has provided a guiding principle, which states that words appearing in similar contexts tend to have similar meanings. This principle has led to a variety of distributional semantic models (DSM) that use multidimensional vectors as word-sense representation. Latent semantic analysis \cite{lsa} is a representative method of this kind, which applies truncated Singular Value Decomposition to a matrix of word-context co-occurrence matrix. Recently, neural network-based methods, such as Skip-Gram \cite{word2vec} and Glove \cite{glove}, have been proposed to represent words as low-dimensional vectors called word embeddings. Compared with traditional DSM methods, these word embeddings have shown superior performance in similarity measurement between words.

Let $\mathbf D\in \mathbb{R}^{d\times |V|}$ be a learned embedding matrix for a finite vocabulary $V$ of $|V|$ words, with $d$ denoting the dimensionality of word embeddings. The $i$-th column, $(\mathbf{D}(i)\in \mathbb{R}^d$, represents the embedding of the $i$-th word in the vocabulary $V$. Given two words $x$ and $y$ in the vocabulary, we can measure their semantic similarity as the most commonly-used cosine similarity between their word embeddings $\mathbf x=\mathbf D(x)$ and $\mathbf y=\mathbf D(y)$:

\begin{align}
sim(x,y)=cosine(\mathbf{x}, \mathbf{y})=\frac{\langle\mathbf{x},\mathbf{y}\rangle}{\|\mathbf{x}\|\cdot\|\mathbf{y}\|}
\end{align}
Therefore, for each word $y_t (1\le t\le m)$ in the hypothesis $Y=y_1y_2...y_m$, we use $\gamma(y_t)$ to denote the word in premise $X=x_1x_2...x_n$ that is of highest semantic similarity with $y_t$:

\begin{align}
\gamma(y_t)=\operatorname*{arg\,max}_{x_s} sim(x_s,y_t)
\end{align}

Such a mapping $\gamma$ can build the connection from the hypothesis to the premise, at the word level. Each sentence pair $(X,Y)$ can then be represented as a sequence $Z=z_1z_2...z_m$ where $z_t=(y_t, \gamma(y_t)), 1\le t\le m,$ denotes the $t$-th word in hypothesis $Y$ paired with its most-similar word $\gamma(y_t)$ in the premise $Y$. This pairing process do associate the most relevant words from the hypothesis $Y$ to the premise $X$. We use $\mathbf{z}_t\in\mathbb{R}^{2d}$ to denote the concatenation of word embeddings of $y_t$ and $\gamma(y_t)$ for $z_t=(y_t, \gamma(y_t))$:

\begin{align}
\mathbf{z}_t = \begin{bmatrix}
				\mathbf{D}(y_t) \\
				\mathbf{D}(\gamma(y_t))\\
			\end{bmatrix}
\end{align}
Thus, we can get a sequence $\mathbf{Z}=(\mathbf{z}_1,\mathbf{z}_2,\dots,\mathbf{z}_m)$ of vectors in $\mathbb{R}^{2d}$,  which can be thought of as an augmented representation of the hypothesis with reference to the premise.

\subsection{Sequence Modeling with LSTM}

Next, we would like to transform the sequence $\mathbf{Z}$ into a vector, as the representation of the hypothesis conditioned on the premise. Recurrent neural networks are naturally suited for modeling variable-length sequences, which can recursively compose each $(2d)$-dimensional vector $\mathbf{z}_t$ with its previous memory. Traditional recurrent neural networks often suffer from the problem of vanishing and exploding gradients \cite{rnn-gradients-1}\cite{rnn-gradients-2}, making it hard to train models. In this paper, we adopt the Long Short-Term Memory (LSTM) model \cite{lstm} which partially solves the problem by using gated activation function.

The LSTM maintains a memory state $\mathbf{c}$ through all the time steps, in order to save the information over long time periods. Concretely, at each time step $t$, the concatenation of two word embeddings, $\mathbf{z}_t$, is fed into the LSTM as the input. The LSTM updates the memory state from the previous $\mathbf{c}_{t-1}$ to the current $\mathbf{c}_t$, by adding new content that should be memorized and erasing old content that should be forgotten. The LSTM also outputs current content that should be exposed. relying on memory is updated by partially forgetting the previous memory $\mathbf{c}_{t-1}$ and adding a new memory content $\tanh (\mathbf{W}^c\mathbf{H}+\mathbf{b}^c)$. The output gates $\mathbf{o}_t$ modulates the amount of memory content exposure.

\begin{align}
\mathbf{H} &= \begin{bmatrix}
				\mathbf{z}_t \\
				\mathbf{h}_{t-1} \\
			\end{bmatrix} \\
\mathbf{i}_t &= \sigma(\mathbf{W}^i\mathbf{H}+\mathbf{b}^i) \\
\mathbf{f}_t &= \sigma(\mathbf{W}^f\mathbf{H}+\mathbf{b}^f) \\
\mathbf{o}_t &= \sigma(\mathbf{W}^o\mathbf{H}+\mathbf{b}^o) \\
\mathbf{c}_t &= \mathbf{f}_t\odot\mathbf{c}_{t-1}+\mathbf{i}_t\odot\tanh(\mathbf{W}^c\mathbf{H}+\mathbf{b}^c) \\
\mathbf{h}_t &= \mathbf{o}_t\odot\tanh(\mathbf{c}_t)
\end{align}

To prevent overfitting, dropout is applied to regularize the LSTMs. Dropout has shown a great success when working with feed-forward networks \cite{dropout}. As indicated in \cite{lstm-dropout}, our method drops the input and the output of the LSTM layer, with the same dropout rate.

\subsection{Decision Making with Softmax Layer}

The final output, $\mathbf{h}_m$, generated by the LSTM on the enhanced representation $\mathbf{Z}$ of the hypothesis $Y$ conditioned on the premise $X$, is fed into a softmax layer which performs the following two steps.

As the first step, $\mathbf{h}_m$ goes through a linear transformation to get a 3-dimensional vector $\mathbf{p}$:
\begin{align}
\mathbf{p} = \mathbf{W}^s\mathbf{h}_m+\mathbf{b}^s
\end{align}
where the weight matrix $\mathbf{W}^s\in\mathbb{R}^{3\times k}$ , and bias vector $\mathbf{b}^s\in\mathbb{R}^3$ are the parameters of the softmax layer.

The $\mathbf{p}=(p_1,p_2,p_3$ is then transformed by a nonlinear softmax function, resulting in a probabilistic prediction $(\hat{t}_1, \hat{t}_2, \hat{t}_3)$ over the three possible labels (Entailment=1, Contradiction=2, or Neutral=3):
\begin{align}
\hat{t}_i=\frac{\exp{p_i}}{\sum_{j=1}^{3} \exp{p_j}}
\end{align}

During the training phase, the cross-entropy error function is used as the cost function. At test time, the label with the highest probability, $\arg\max_{1\le i\le 3} p_i$, is output as the predicted label.

\section{Improvements over the Base Method}\label{sec:improvements}

To obtain better predictive performance, three optional techniques are applied on the base method described above:
\begin{itemize}
\item Multiple word-embedding libraries, which improves the vector representations of words;
\item Biway-LSTM integration, which enhances the representations of relations between text pairs;
\item Ensemble based on model averaging, which produces more accurate predictions.
\end{itemize}

\subsection{Mutliple Embeddings}

Word2vec and Glove are two popular software for learning word embeddings from text corpus. We use $\mathbf{D}_{w2v}$ and $\mathbf{D}_{glove}$ to denote their induced word embedding libraries, respectively. For each word $x$, we can represent it as a vector $\mathbf{D}(x)$ as the concatenation of its embedding from $\mathbf{D}_{w2v}$ and that from $\mathbf{D}_{glove}$, that is, $$\mathbf{D}(x)=[\mathbf{D}_{w2v}(x) \mathbf{D}_{glove}(x)]$$.

Or equivalently, a new embedding matrix $\mathbf{D}\in\mathbb{R}^{(2d)\times |V|}$ is constructed by concatenating $\mathbf{D}_{w2v}$ with $\mathbf{D}_{glove}$. The semantic similarity between words from the hypothesis and the premise is calculated with regards to this new embedding matrix. The aim of using this technique is to integrated the potentially complementary information provided by different word embedding libraries. As another  possible solution, we can also make use of canonical correlation analysis (CCA) to project the two embedding libraries into a common semantic space, and thus induce a new embedding library. However, we do not make any further exploration, because it is out of the scope of this paper.

\subsection{Biway-LSTM Integration}
In RTE problem, the premise should also play an important role as the hypothesis. In the previous two subsections, we have described how to model the hypothesis conditioned on the premise. This idea can also be applied the other way round, i.e. to model the premise conditioned on the hypothesis. 

To justify this statement, let us consider the following simple example. Let $X^{(1)}$=``John failed to pass the exam.'', $X^{(2)}$=``John succeeded in passing the exam.'', and $Y$=``John passed the exam.'' It is clear that $X^{(2)}$ entails $Y$, but $X^{(1)}$ contradicts $Y$. However, the enhanced representation of $Y$ conditioned on $X^{(1)}$ is the same as that of $Y$ conditioned on $X^{(2)}$. Therefore, we cannot discriminate these two situations based on only the enhanced representation of $Y$. 

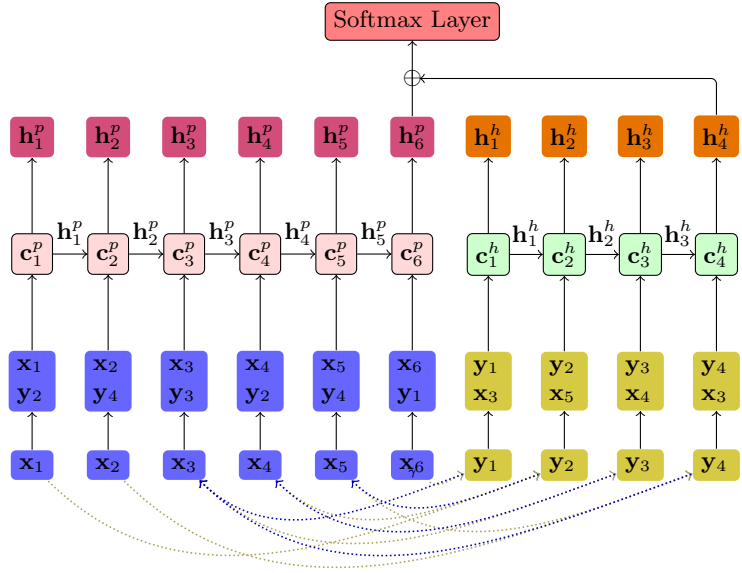
\begin{figure}
\centering
\begin{tikzpicture}[rounded corners=2pt, scale=.5]
   \pgfmathsetseed{1}
    \foreach \t in {-5,...,0} {
    \pgfmathtruncatemacro{\addsix}{\t + 6};
    \pgfmathtruncatemacro{\hypo}{random(1, 4)};
    \node[rectangle, fill=blue!60!white] (prem\addsix) at (\t * 2, 0)
                           {$\mathbf{x}_\addsix$};
    \node[rectangle, fill=blue!60!white, text width=.4cm] (premconcat\addsix) [above=.5cm of prem\addsix]
                               {$\mathbf{x}_\addsix$ $\mathbf{y}_\hypo$};
    \path[->] (prem\addsix) edge (premconcat\addsix);
   }
   
   \pgfmathsetseed{1}
    \foreach \t in {1,...,4} {
    \pgfmathtruncatemacro{\prem}{-random(0, 5)};
    \pgfmathtruncatemacro{\addsix}{\prem + 6};   
    \node[rectangle, fill=yellow!80!black, text width=.4cm] (hypo\t) at (\t * 2, 0)
                          {$\mathbf{y}_\t$};
    \node[rectangle, fill=yellow!80!black, text width=.4cm] (hypoconcat\t) [above=.5cm of hypo\t]
                              {$\mathbf{y}_\t$ $\mathbf{x}_\addsix$};
    \path[->] (hypo\t) edge (hypoconcat\t);
   }
   
   \pgfmathsetseed{1}
    \foreach \t in {-5,...,0} {
    \pgfmathtruncatemacro{\addsix}{\t + 6};
    \pgfmathtruncatemacro{\hypo}{random(1, 4)};
    \draw[densely dotted, <-,semithick, yellow!60!black] (hypo\hypo) edge[out=200, in=315] (prem\addsix)
         node[near end,sloped,below] {\small{$\gamma$}};
   	\node[draw, rectangle, fill=pink!60] (pstep\addsix) [above=of premconcat\addsix] {$\mathbf{c}^p_\addsix$};
   	\node[rectangle, fill=purple!70!white] (phid\addsix) [above=of pstep\addsix] {$\mathbf{h}^p_\addsix$};
   	\path[->] (premconcat\addsix) edge (pstep\addsix);
   	\path[->] (pstep\addsix) edge (phid\addsix);
   }
   \foreach \t in {-5,...,-1} {
   \pgfmathtruncatemacro{\addsix}{\t + 6};
    \pgfmathtruncatemacro{\next}{\addsix + 1};
   	\path[->] (pstep\addsix) edge node[above]{$\mathbf{h}^p_\addsix$} (pstep\next);
   }
   
   \pgfmathsetseed{1}
    \foreach \t in {1,...,4} {
    \pgfmathtruncatemacro{\prem}{-random(0, 5)};
    \pgfmathtruncatemacro{\addsix}{\prem + 6};
    \draw[densely dotted, <-,semithick, blue!70!black] (prem\addsix) edge[out=315, in=200] (hypo\t)
      node[near end,sloped,below] {\small{$\gamma$}};
	\node[draw, rectangle, fill=green!20] (step\t) [above=of hypoconcat\t] {$\mathbf{c}^h_\t$};
	\node[rectangle, fill=orange!90!black] (hid\t) [above=of step\t] {$\mathbf{h}^h_\t$};
	\path[->] (hypoconcat\t) edge (step\t);
	\path[->] (step\t) edge (hid\t);
  }
  \foreach \t in {1,...,3} {
  	\pgfmathtruncatemacro{\next}{\t + 1};
	\path[->] (step\t) edge node[above]{$\mathbf{h}^h_\t$} (step\next);
  }
  
  \node[draw, rectangle, fill=red!50] (softmax) [above=of phid6] {Softmax Layer};
  \path[->] (phid6) edge node[inner sep=-1pt](concat){$\oplus$} (softmax);
  \path[->, to path={|- (\tikztotarget)}] (hid4) edge (concat);

\end{tikzpicture}

\caption{The Biway Architecture of MaxCosine-LSTM}
\label{fig:biway-maxcosinelstm}
\end{figure}

To do a remedy, we extend our base model to the biway architecture illustrated in \Cref{fig:biway-maxcosinelstm}. Two LSTMs are used to separately model the enhanced representation of the premise and that of the hypothesis. Their final output vectors are then concatenated and fed into a softmax layer to do the final decision.

\subsection{Ensemble by Model Averaging}
Combining multiple models generally leads to better performance. The component models are expected to be diverse and accurate, in order to produce an ensemble of high quality. In this paper, all the components are homogeneous, that is, all of them are induced by our base method (optionally enhanced with bi-embedding integration and/or biway integration). The diversity of these components comes from random initialization of the model parameters, with different random seeds. The predictions from these component models are averaged to make the final decision.

\section{Experiments}\label{sec:experiment}

In the experimental part, we evaluate our method on the Stanford Natural Language Inference (SNLI) dataset \cite{snli-dataset} which consists of about 570K sentence pairs. After filtering sentence pairs with unknown class labels, we get a train data of 549,367 pairs, a validation data of 9,842 pairs, a test data of 9,824 pairs. This dataset has been commonly used by previous state-of-the-art neural models. 

To train our model, we use cross-entropy loss $J(\theta)$ in Equation \eqref{eq:cost-func},
the $B$ is the mini-batch size, $t^{(i)}$ is the true label of sample $i$, $\{t^{(i)} = j\}$ is
$1$ if $t^{(i)}$ equals $j$ else $0$.
We use stochastic mini-batch gradient descent with the ADAM optimizer\cite{adam}, we set ADAM's
hyperparameters $\beta_1 = 0.9$ and $beta_2 = 0.999$ and the initial learning rate to 0.001.
We use both the pre-trained Glove\cite{glove} model \textbf{glove.840B.300d} and
Word2vec \cite{word2vec} model \textbf{GoogleNews-vectors-negative300} to initialize word embeddings.
We don't tune the word embeddings when train, OOV words' vectors are set to be the average
of their window words' vectors, follow same setting in the Match-LSTM paper\cite{match-lstm}.
We fix the length of LSTM hidden states $k$ to 300D, and apply various dropout rate on the input layer.
We don't apply any regularization to the network weights, and use batch size $B=128$ when training.
We use Lasagne\footnote{http://lasagne.readthedocs.io/en/latest/} to implement these models.
\begin{align} \label{eq:cost-func}
J(\theta) = - \frac{1}{B} \sum_{i=1}^{B} \sum_{j=1}^{3}  \mathbf{1}\{t^{(i)} = j\} \log \hat{t}_j
\end{align}

We compare our approach with the following state-of-art methods:
\begin{itemize}
\item Separate-LSTM: the first neural method proposed in \cite{snli-dataset}, which encodes the premise and the hypothesis with two separate LSTMs independently. 
\item Sequential-LSTM method: this method \cite{attention-lstm} makes use of two LSTMs, where an LSTM reads the premise, and a second LSTM initialized with the final cell state of the first LSTM reads a delimiter and the hypothesis.
\item Attention-LSTM: the method in \cite{attention-lstm} that attends over output vectors of the premise only for the final output of the hypothesis.
\item Word-by-Word Attention-LSTM: the method in \cite{attention-lstm} that attends over output vectors of the premise for every word in the hypothesis.
\item matchLSTM with word embedding: the method in \cite{match-lstm} that performs
word-by-word matching of the hypothesis with the premise.
\end{itemize}
We implemented our versions of Attention-LSTM, Word-by-Word Attention-LSTM, and matchLSTM, because the codes of the original papers are not made publicly available.

Results on the SNLI corpus are summarized in Table \ref{tab:comparison-table}. Our method MaxCosine-LSTM that uses all three improvement techniques has achieved the best performance when compared with these state-of-the-art methods.

\begin{table}
\caption{Empirical accuracy of the MaxCosine-LSTM model compared with previous results}
\label{tab:comparison-table}
\begin{center}
\begin{tabular}{||m{7.8cm}|m{0.8cm}|m{2.5cm}||}
    \hline
	\bf Model	                                  & $\mathbf k$ & \bf Test Accuracy \\ \hline
	Separate-LSTM      							  & 100   & 77.8     \\
	Sequential-LSTM                               & 100   & 80.9     \\
	Attention-LSTM								  & 100   & 82.3     \\
	Word-by-Word Attention-LSTM (our implementation)   & 100   & 83.3     \\
	mLSTM with word embedding(our implementation) & 300   & 84.2    \\ \hline
	MaxCosine-LSTM-biEmb-biWay-Ensemble           & 300   & \bf{85.0}    \\\hline
\end{tabular}
\end{center}
\end{table}

We also conduct experiments to study the effectiveness of the three techniques in \Cref{sec:improvements}. With the dropout ratio set as 0.3, 0.4, and 0.5, \Cref{tab:results-table} shows the results of using different combinations of the three techniques. The accuracies of all the non-ensemble methods (MaxCosine-LSTM, MaxCosine-LSTM-biEmb, MaxCosine-LSTM-biWay, MaxCosine-LSTM-biEmb-biWay) are reported as the average over 5 runs.

It can be easily observed that all the three techniques have their own contributions to the final predictive ability. Discarding any technique would lead to some decrease in the prediction accuracy. It is also evident that the ensemble technique is the most important one, the biway integration goes next, and the bi-embedding integration is relatively less significant.

In addition, we can also observe the following two facts:
\begin{itemize}
\item The effect of ensemble technique on MaxCosine-LSTM-biWay is more significant than its effect on MaxCosine-LSTM or MaxCosine-LSTM-biEmb.
\item The effect of ensemble is more significant with smaller dropout ratio.
\end{itemize}
These observations can be explained by the fact that the ensemble technique based on model averaging works better on the method of higher model complexity which usually has higher variance in bias-variance decomposition. Intuitively, the model complexity of MaxCosine-LSTM-biWay is much higher than that of MaxCosine-LSTM and that of MaxCosine-LSTM-biEmb. Another fact is that the dropout ratio can control the model complexity: lower dropout ratio means higher model complexity, because dropout is one kind of regularization. 

\begin{table}
\caption{Effects of the three techniques for MaxCosine-LSTM model}
\label{tab:results-table}
\begin{center}
\begin{tabular}{||m{1.4cm}|m{6.5cm}|m{1.6cm}||}
    \hline
	\textbf{Dropout Ratio}   	& \textbf{Techiques Used}    	& \textbf{Accuracy} \\
	\hline
	\multirow{8}{1cm}{0.3}   	& MaxCosine-LSTM  				& 82.45\%  \\
	     						& MaxCosine-LSTM-biEmb 			& 82.50\%  \\
	     						& MaxCosine-LSTM-biWay			& 82.93\%  \\
	     						& MaxCosine-LSTM-Ensemble     	& 83.65\%  \\
	     						& MaxCosine-LSTM-biEmb-biWay  	& 83.26\%  \\
	     						& MaxCosine-LSTM-biEmb-Ensemble & 83.92\%  \\
	     						& MaxCosine-LSTM-biWay-Ensemble & 84.78\%  \\
	     						& MaxCosine-LSTM-biEmb-biWay-Ensemble  	& 84.98\%  \\
	\hline
	\multirow{8}{1cm}{0.4}   	& MaxCosine-LSTM  				& 82.07\%  \\
	     						& MaxCosine-LSTM-biEmb 			& 82.56\%  \\
	     						& MaxCosine-LSTM-biWay			& 83.05\%  \\
	     						& MaxCosine-LSTM-Ensemble     	& 83.09\%  \\
	     						& MaxCosine-LSTM-biEmb-biWay  	& 83.31\%  \\
	     						& MaxCosine-LSTM-biEmb-Ensemble & 83.57\%  \\
	     						& MaxCosine-LSTM-biWay-Ensemble & 84.23\%  \\
	     						& MaxCosine-LSTM-biEmb-biWay-Ensemble  	& 84.74\%  \\
 	\hline	
	\multirow{8}{1cm}{0.5}   	& MaxCosine-LSTM  				& 81.68\%  \\
	     						& MaxCosine-LSTM-biEmb 			& 82.18\%  \\
	     						& MaxCosine-LSTM-biWay			& 82.77\%  \\
	     						& MaxCosine-LSTM-Ensemble     	& 82.37\%  \\
	     						& MaxCosine-LSTM-biEmb-biWay  	& 83.44\%  \\
	     						& MaxCosine-LSTM-biEmb-Ensemble & 83.05\%  \\
	     						& MaxCosine-LSTM-biWay-Ensemble & 83.89\%  \\
	     						& MaxCosine-LSTM-biEmb-biWay-Ensemble  	& 84.43\%  \\
	\hline

\end{tabular}
\end{center}
\end{table}

\section{Conclusion and Outlook}
We proposed a simple neural method to determine the relation between a hypothesis and a premise. It relies on word semantic matching from the hypothesis to the premise (or vice versa), then makes use the LSTM to do the sequence modeling, and finally feed the final output from the LSTM into a softmax layer to make the classification decision. After equipped with three techniques to improve its performance, experimental results have shown that our method has achieved better accuracies than state-of-the-art systems. In addition, it is also shown that the three techniques all have their own contribution to the accuracy obtained.

\section*{Acknowledgments}

This work is supported by National High-Tech R\&D Program of China
(863 Program) (No. 2015AA015404), and Science and Technology Commission of Shanghai Municipality (No. 14511106802). We are grateful to the anonymous reviewers for their valuable comments.

\end{document}